\documentclass[11pt,a4paper]{article}
\usepackage[hyperref]{acl2021}
\usepackage{times}
\usepackage{latexsym}

\usepackage{microtype}
\usepackage{graphicx}
\usepackage{url}
\usepackage{hyperref}
\usepackage{breakurl}
\usepackage{array, multirow}
\usepackage{tabularx, lipsum}
\usepackage{float}
\usepackage{bm}
\usepackage{amsmath}

\def\thick{\noalign{\hrule height 1pt}}
\newcolumntype{s}{>{\hsize=.45\hsize}X}

\aclfinalcopy 
\def\aclpaperid{1742} 

\title{On the Interplay Between Fine-tuning and Composition in Transformers}

\author{Lang Yu \\
  Deptartment of Computer Science \\
  University of Chicago \\
  \texttt{langyu@uchicago.edu} \\\And
  Allyson Ettinger \\
  Department of Linguistics \\
  University of Chicago \\
  \texttt{aettinger@uchicago.edu} \\}

\date{}

\begin{document}
\maketitle
\begin{abstract}
Pre-trained transformer language models have shown remarkable performance on a variety of NLP tasks. However, recent research has suggested that phrase-level representations in these models reflect heavy influences of lexical content, but lack evidence of sophisticated, compositional phrase information~\cite{yu2020assessing}. Here we investigate the impact of fine-tuning on the capacity of contextualized embeddings to capture phrase meaning information beyond lexical content. Specifically, we fine-tune models on an adversarial paraphrase classification task with high lexical overlap, and on a sentiment classification task. After fine-tuning, we analyze phrasal representations in controlled settings following prior work. We find that fine-tuning largely fails to benefit compositionality in these representations, though training on sentiment yields a small, localized benefit for certain models. In follow-up analyses, we identify confounding cues in the paraphrase dataset that may explain the lack of composition benefits from that task, and we discuss potential factors underlying the localized benefits from sentiment training. 
\end{abstract}

\section{Introduction}

Transformer language models like BERT \cite{devlin2019bert}, GPT \cite{radford2018improving, radford2019language} and XLNet \cite{yang2019xlnet}, have improved the state-of-art in many NLP tasks since their introduction. The versatility of these pre-trained models suggests that they may acquire fairly robust linguistic knowledge and capacity for natural language ``understanding''. However, an emerging body of analysis demonstrates a level of superficiality in these models' handling of language \cite{niven2019probing, kim2020cogs, mccoy2019right, ettinger2020bert,yu2020assessing}.


In particular, although \emph{composition}---a model's capacity to combine meaning units into more complex units reflecting phrase meanings---is an indispensable component of language understanding, when testing for composition in pre-trained transformer representations, \citet{yu2020assessing} report that these representations reflect word content of phrases, but don't show signs of more sophisticated humanlike composition beyond word content. In the present paper we perform a direct follow-up of that study, asking whether models will show better evidence of composition after fine-tuning on tasks that are good candidates for requiring composition: 1) the Quora Question Pairs dataset in Paraphrase Adversaries from Word Scrambling (PAWS-QQP) \cite{zhang2019paws}, an adversarial paraphrase dataset forcing models to classify paraphrases with high lexical overlap, and 2) the Stanford Sentiment Treebank \cite{socher2013recursive}, a sentiment dataset with fine-grained phrase labels to promote composition. We base our analysis on the tests proposed by \citet{yu2020assessing}, which rely on alignment with human judgments of phrase pair similarities, and which leverage control of lexical overlap to target compositionality. We fine-tune and evaluate the same models and representation types tested in that paper, for optimal comparison.


We find that across the board, fine-tuning on PAWS-QQP does not improve compositionality---if anything, performance on composition metrics tends to degrade. Composition performance also remains low after training on SST, but we do see some localized improvements for certain models. Analyzing the PAWS-QQP dataset, we find reliable superficial cues to paraphrase labels (distance of word swap), explaining in part why fine-tuning on that task might fail to improve composition---and reinforcing the need for caution in interpreting difficulty of NLP tasks. We also discuss the contribution of variation in size of labeled phrases in SST, with respect to the benefits that result from fine-tuning on that task. All experimental code and data are made available for further testing.\footnote{Datasets and code available at https://github.com/yulang/fine-tuning-and-composition-in-transformers}

\begin{table*}[ht]
    \centering
    \begin{tabularx}{\linewidth}{XXc}
    \thick
\textbf{Sentence 1}         & \textbf{Sentence 2}  & \textbf{Label}      \\ \thick
There are also specific discussions , public profile debates and project discussions .
 & There are also public discussions , profile specific discussions , and project discussions .
 & 0 \\ \cline{1-3}
She worked and lived in Stuttgart , Berlin ( Germany ) and in Vienna ( Austria ) .
& She worked and lived in Germany ( Stuttgart , Berlin ) and in Vienna ( Austria ) .
& 1          \\ \thick
    \end{tabularx}
\caption{Example pairs from PAWS-QQP. Both positive and negative pairs have high bag-of-words overlap.}
\label{tab:paws_example}
\end{table*}

\section{Related work}

Extensive work has studied the nature of learned representations in NLP models~\cite{adi2016fine, conneau2018you, ettinger2016probing, durrani2020analyzing}. Our work builds in particular on analysis of contextualized representations \cite{bacon2019does, tenney2019you,peters2018dissecting,hewitt2019structural,klafka2020spying, toshniwal2020cross}. Other work that has focused on transformers, as we do, has often focused on analyzing the attention mechanism \cite{vig2019analyzing, clark2019does}, learned parameters \cite{roberts2020much, radford2019language, raffel2020exploring} and redundancy \cite{dalvi2020analyzing, voita2019analyzing, michel2019sixteen}. The evaluation that we use here follows the paradigm of classification-based probing~\cite{kim2019probing, wang2018glue, paws2019naacl, pawsx2019emnlp} and correlation with similarity judgments \cite{finkelstein2001placing, gerz2016simverb, hill2015simlex, conneau2018senteval}. 

The current paper also builds on work subjecting trained NLP models to adversarial inputs, to highlight model weaknesses. One body of work approaches the problem by applying heuristic rules of perturbation to input sequences \cite{wallace2019universal, jia2017adversarial, zhang2019paws}, while another uses neural models to construct adversarial examples~\cite{li2020bert, li2018textbugger} or manipulate inputs in embedding space~\cite{jin2020bert}. Our work also contributes to efforts to understand impacts and outcomes of the fine-tuning process~\cite{miaschi2020linguistic, mosbach2020interplay, wang2020meta, perezmayos2021evolution}.

Phrase and sentence composition has drawn frequent attention in analysis of neural models, often focusing on analysis of internal representations and downstream task behavior~\cite{ettinger2018assessing, conneau2019unsupervised, nandakumar2019well, mccoy2019right, yu2020assessing, bhathena2020evaluating, mu2020compositional, andreas2019measuring}. Some work investigates compositionality via constructing linguistic~\cite{keysers2019measuring} and non-linguistic~\cite{livska2018memorize, hupkes2018learning, baan2019realization} synthetic datasets.

Most related to our work here is the finding of~\citet{yu2020assessing}. They test for composition in two-word phrase representations from transformers, via similarity correlations and paraphrase detection. They find that baseline performance on these tasks is high, but once they control for amount of word overlap, performance drops dramatically, suggesting that observed correspondences rely on word content rather than phrase composition. We build directly on this work, testing whether these patterns will still hold after fine-tuning on tasks intended to encourage composition.

\section{Fine-tuning Pre-trained Transformers}\label{sec:fine-tune}

In response to the weaknesses observed by~\citet{yu2020assessing}, we select two different datasets with promising characteristics for addressing these weaknesses. We fine-tune on these tasks, then perform layer-wise testing on contextualized representations from the fine-tuned models, comparing against results on the pre-trained models. Here we describe the two fine-tuning datasets.

\subsection{PAWS: fine-tuning on high word overlap}

The core of the~\citet{yu2020assessing} finding is that model performance on the selected composition tests degrades significantly when cues of lexical overlap are controlled. It stands to reason, then, that a model trained to discern meaning differences under conditions of high lexical overlap may improve on these overlap-controlled composition tests. This drives our selection of the Paraphrase Adversaries from Word Scrambling (PAWS) dataset~\cite{paws2019naacl}, which consists of sentence pairs with high lexical overlap. The task is formulated as binary classification of whether two sentences are paraphrases or not. State-of-the-art models achieve only $<40$\% accuracy before training on the dataset~\cite{zhang2019paws}. Table~\ref{tab:paws_example} shows examples from this dataset. Due to the high lexical overlap, we might expect that in order to achieve non-trivial accuracy on this task, models must attend to more sophisticated meaning information than simple word content.

\subsection{SST: fine-tuning on hierarchical labels}

Another dataset that has been associated with training and evaluation of phrasal composition is the Stanford Sentiment Treebank, which contains syntactic phrases of various lengths, together with fine-grained human-annotated sentiment labels for these phrases. Because this dataset contains annotations of composed phrases of various sizes, we can reasonably expect that training on this dataset may foster an increased sensitivity to compositional phrase meaning. We formulate the fine-tuning task as a 5-class classification task following the setup in~\citet{socher2013recursive}.
The models are trained to predict sentiment labels given phrases as input.

\section{Representation evaluation}
\label{rep_com_eval}

For optimal comparison of the effects of fine-tuning on the above tasks, we replicate the tests, representation types, and models reported on by \citeauthor{yu2020assessing}. Here we briefly describe these methods. For more details on the evaluation dataset and task setup, please refer to~\citet{yu2020assessing}.

\subsection{Evaluation tasks}
\citeauthor{yu2020assessing} propose two analyses for measuring composition: similarity correlations and paraphrase classification. They focus on two-word phrases, using the BiRD bigram relatedness dataset \cite{asaadi2019big} for similarity correlations, 
and the PPDB 2.0 paraphrase database \cite{ganitkevitch2013ppdb, pavlick2015ppdb} for paraphrase classification. BiRD contains 3,345 bigram pairs, with source phrases paired with numerous target phrases, and human-annotated similarity scores ranging from 0 to 1. For similarity correlation, \citeauthor{yu2020assessing} take layer-wise correlations between these human phrase similarity scores and the cosine similarities of model representations for the same phrases. For paraphrase classification, \citeauthor{yu2020assessing} train a multi-layer perceptron classifier to label whether two phrase representations are paraphrases, drawing their positive phrase pairs from PPDB 2.0---which contains paraphrases with scores generated by a regression model---and randomly sampling negative pairs from the rest of the dataset. We replicate all of these procedures.

For both task types,~\citeauthor{yu2020assessing} compare between ``uncontrolled''  and ``controlled'' tests, with the latter filtering the data to control word overlap within phrase pairs, such that amount of word overlap between two phrases can no longer be used as a cue for how similar the meanings are. 
It is on these controlled settings that \citeauthor{yu2020assessing} observe the significant drop in performance, suggesting that model representations lack the compositional knowledge to discern phrase meaning beyond word content. Below we will report results for both settings, with particular focus on controlled settings.

\begin{figure*}[ht]
    \centering
    \includegraphics[width=1\textwidth]{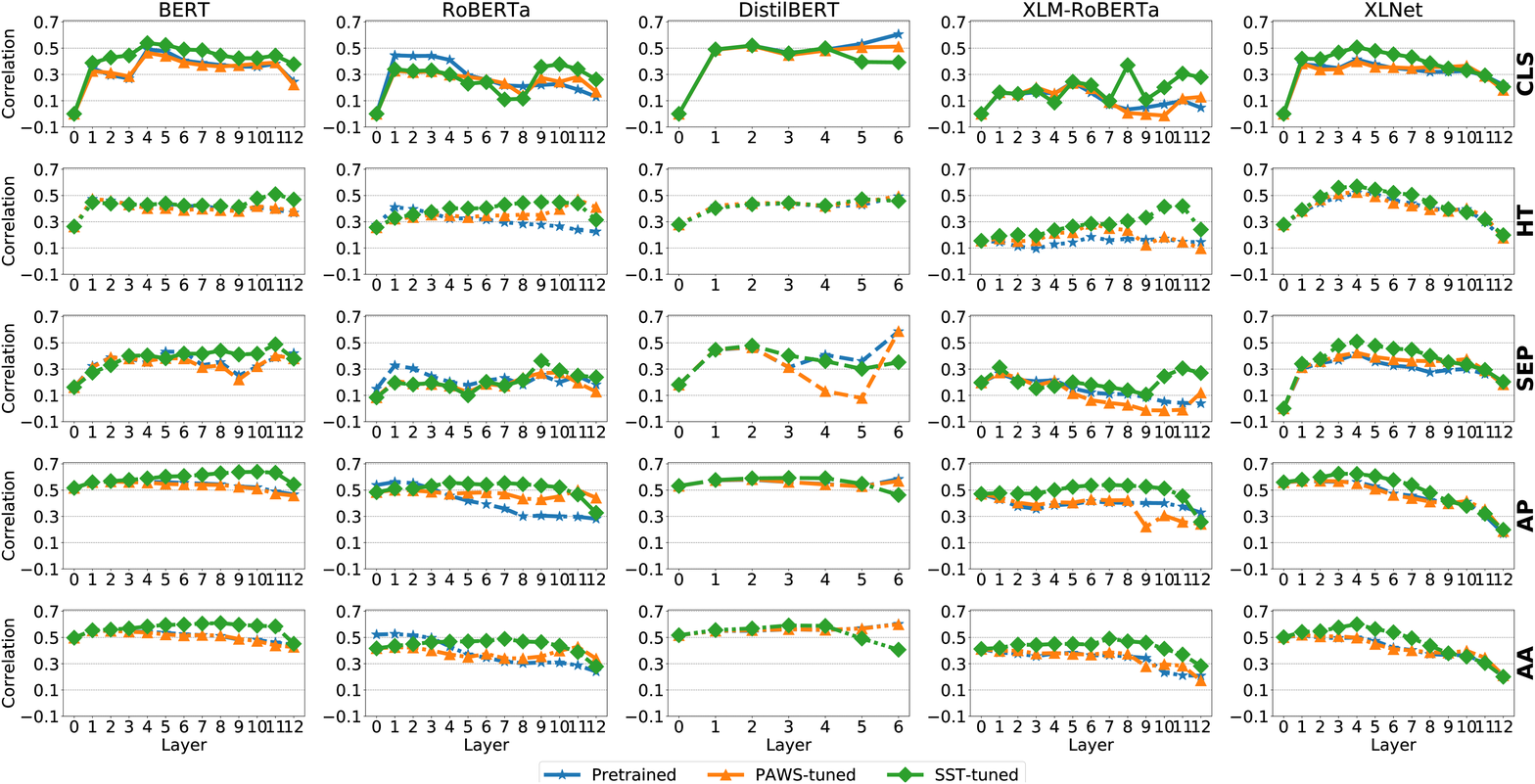}
    \caption{Similarity correlation on uncontrolled BiRD dataset, with phrase-only input. Columns correspond to models, and rows correspond to representation types (``HT'' = Head-token, ``AP'' = Avg-Phrase and ``AA'' = Avg-All). For each model and representation type, the corresponding subplot shows correlations for pre-trained, PAWS-tuned and SST-tuned settings, respectively. For each subplot, X-axis corresponds to layer index, and Y-axis corresponds to correlation value. Layer 0 corresponds to input embeddings passed to the model.}
    \label{fig:bird-tuned}
\end{figure*}

\subsection{Representation types}

Following~\citeauthor{yu2020assessing}, for each input phrase we test as a potential representation 1) 
CLS token, 2) average of tokens within the phrase (Avg-Phrase), 3) average of all input tokens (Avg-All), 4) embedding of the second word of the phrase, intended to approximate the semantic head (Head-Word), and 5) SEP token. We test each of these representations at every layer of each model.\footnote{Like~\citeauthor{yu2020assessing}, we also test both phrase-only input (encoder input consists only of two-word phrase plus special CLS/SEP tokens), as well as inputs in which phrases are embedded in sentence contexts.}

\begin{figure*}[ht]
    \centering
    \includegraphics[width=1\textwidth]{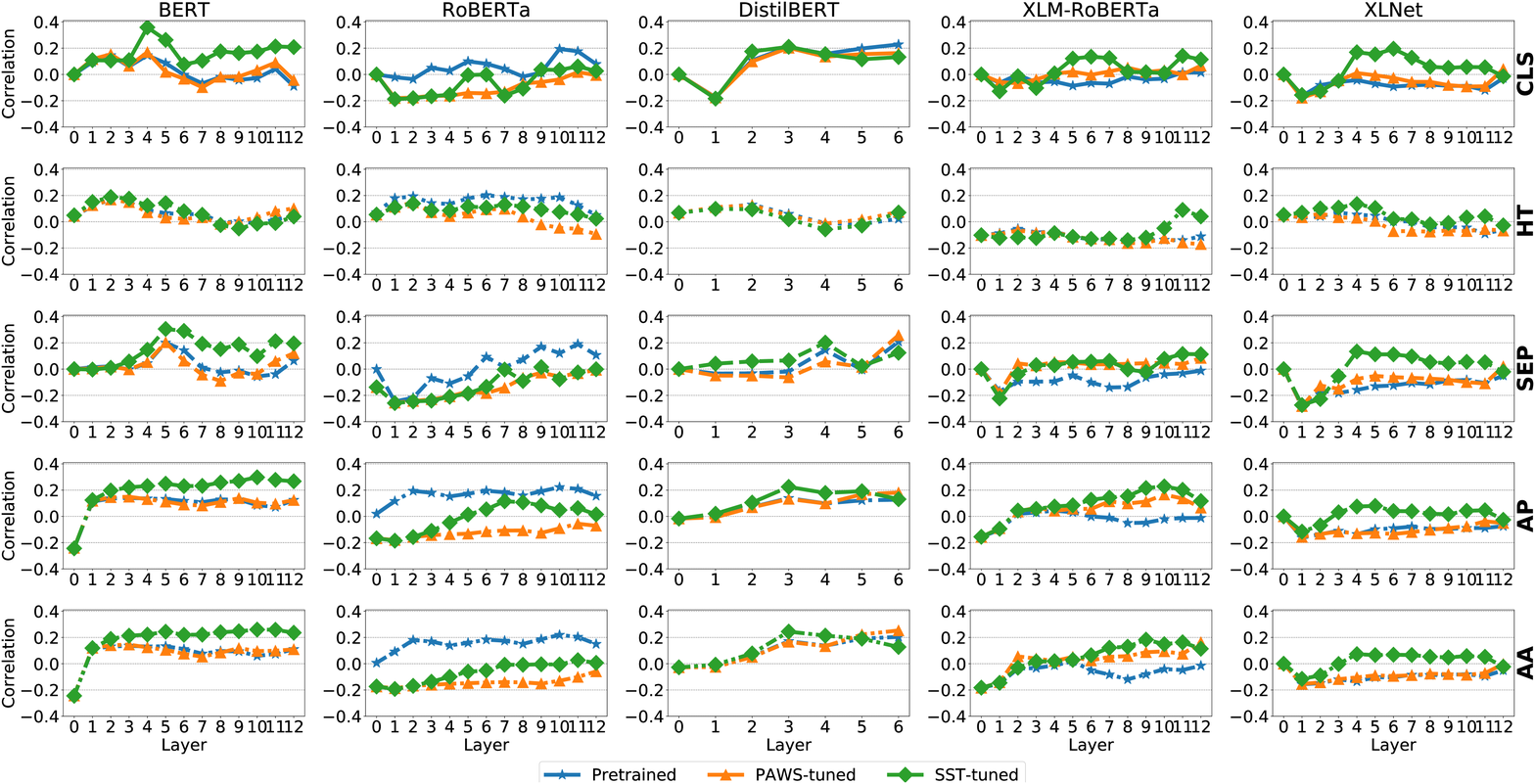}
    \caption{Similarity correlation on controlled BiRD dataset (AB-BA setting), with phrase-only input.}
    \label{fig:bird-abba-tuned}
\end{figure*}

\begin{figure*}[ht]
    \centering
    \includegraphics[width=1\textwidth]{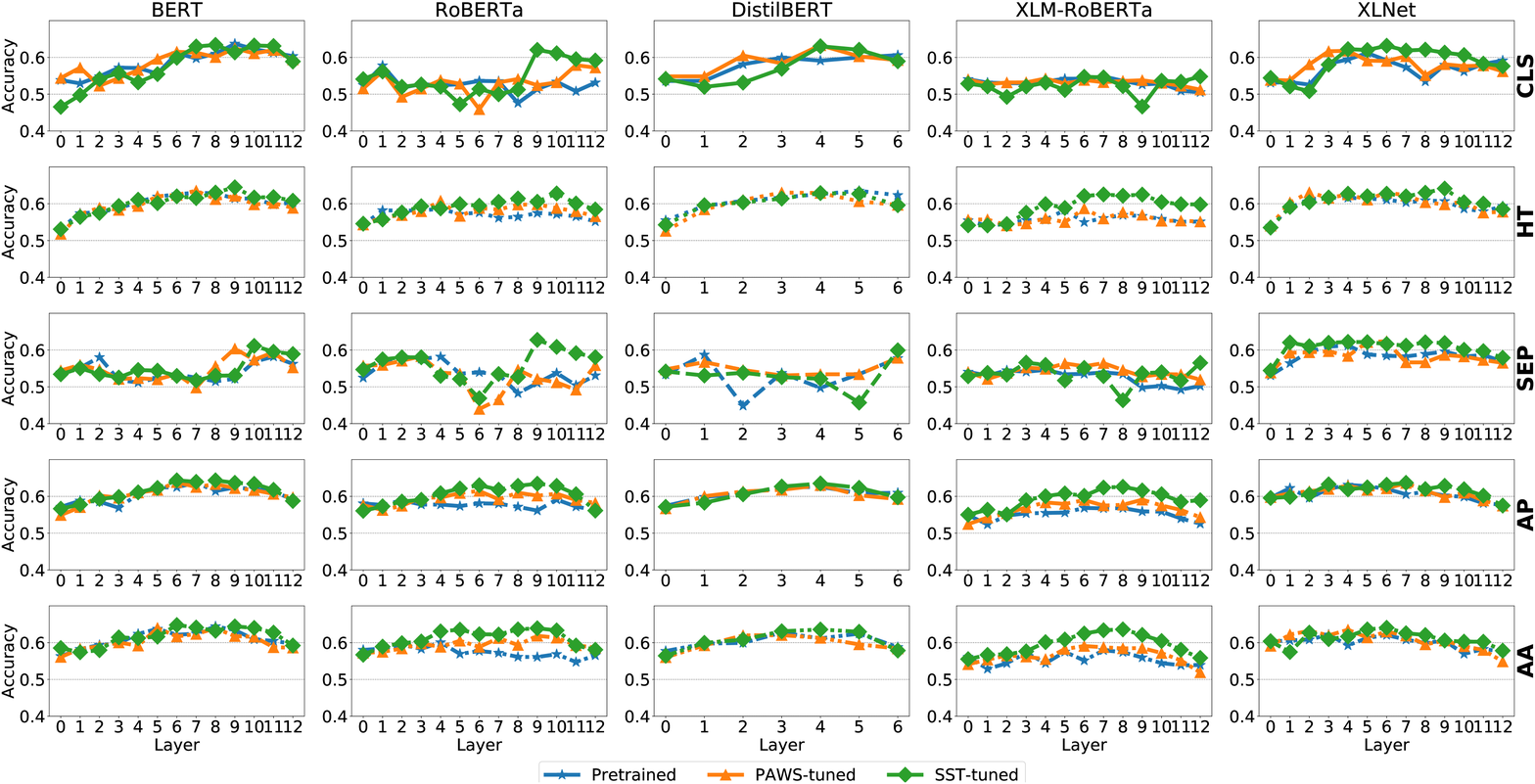}
    \caption{Paraphrase classification accuracy on controlled PPDB dataset (50\% word overlap setting) with phrase-only input. Y-axis range is smaller relative to Figure~\ref{fig:ppdb-tuned}, to make changes from pre-training more visible. 
    }
    \label{fig:ppdb-exact-tuned}
\end{figure*}

\section{Experimental setup}

We fine-tune and analyze the same models that~\citeauthor{yu2020assessing} test in pre-trained form: BERT \cite{devlin2019bert}, RoBERTa \cite{liu2019roberta}, DistilBERT \cite{sanh2019distilbert}, XLNet \cite{yang2019xlnet} and XLM-RoBERTa \cite{conneau2019unsupervised}. In each case, the pre-trained ``base'' version is used as the starting point for fine-tuning. We use the implementation of \citet{Wolf2019HuggingFacesTS}\footnote{\url{https://github.com/huggingface/transformers}} based on PyTorch \cite{paszke2019pytorch}. 

We fine-tune these models on the two datasets described in Section~\ref{sec:fine-tune}. The Quora Question Pairs dataset in Paraphrase Adversaries from Word Scrambling (PAWS-QQP)\footnote{\url{https://github.com/google-research-datasets/paws}} consists of a training set with 11,988 sentence pairs, and a dev/test set with 677 sentence pairs. Tuning on PAWS-QQP is formulated as binary classification. Sentences are passed as input and models are trained to predict whether the input sentences are paraphrases or not. Models are trained on the training set, and validated on the dev/test set for convergence. 

The Stanford Sentiment Treebank (SST)\footnote{\url{https://nlp.stanford.edu/sentiment/treebank.html}} \cite{socher2013recursive} contains 215,154 phrases. 15\% of the data is reserved for validation. The fine-tuning task is formulated as 5-class classification on sentiment labels, where models are given phrases as input, and asked to predict sentiment. In both tasks, we use the Adam optimizer \cite{kingma2014adam} with default weight decay. We train the models until convergence on the validation set. 
 
The evaluation tasks consist of correlation analysis and paraphrase classification. For correlation in the uncontrolled setting, we use the complete BiRD dataset, containing 3,345 phrase pairs.\footnote{\url{http://saifmohammad.com/WebPages/BiRD.html}} For the controlled test, we filter the complete dataset following the criteria in~\citet{yu2020assessing}, resulting in 410 ``AB-BA'' mirror-image pairs with 100\% word overlap (e.g., \emph{law school} / \emph{school law}). For the classification tasks, we use the preprocessed data released by \citet{yu2020assessing}.\footnote{\url{https://github.com/yulang/phrasal-composition-in-transformers}} We collect 12,036 source-target phrase pairs from the preprocessed dataset for our uncontrolled classification setting, and for the controlled classification setting, we collect 11,772 phrase pairs with exactly 50\% word overlap in each pair, following the procedure from the original paper. 

\section{Results after fine-tuning}\label{analysis}

\subsection{Full datasets} Figure \ref{fig:bird-tuned} presents the original results from \citet{yu2020assessing} on pre-trained models, alongside our new results after fine-tuning, on the full BiRD dataset. Since this is prior to the control of word overlap, these correlations can be expected to reflect effects of lexical content encoding, without yet having isolated effects of composition. We find that after fine-tuning on SST, most models and representation types show small improvements in peak correlations across layers, while fine-tuning on PAWS also yields improvements in peak correlations---albeit even smaller---in models other than BERT and XLM-RoBERTa. Overall, within a given representation type, improvements are generally stronger after fine-tuning on SST than on PAWS. Between representation types, Avg-Phrase and Avg-All remain consistently at the highest correlations after fine-tuning. Additionally, we see that the decline in correlation at later layers in pre-trained BERT, RoBERTa and XLM-RoBERTa is mitigated after fine-tuning. Model-wise, we see the most significant improvements in the RoBERTa model, for which the correlations become more consistent across layers for most representation types. As we discuss below, we take this as indication that the fine-tuning promotes more robust retention of word content information across layers, if not more robust phrasal composition.

For the sake of space, we present the plots of the uncontrolled paraphrase classification setting in Figure~\ref{fig:ppdb-tuned} of the Appendix. The overall improvements are even smaller than those seen in the correlations, but we do see comparable patterns in these paraphrase classification results, in particular with SST showing slightly stronger benefits than PAWS.

\subsection{Controlled datasets} Above we see small benefits of fine-tuning for performance on the full, uncontrolled datasets. However, the critical question for our purposes is whether correlations also show improvements in word-overlap controlled settings, which better isolate effects of composition.  
Figure \ref{fig:bird-abba-tuned} shows correlations for all models on the controlled AB-BA (full word overlap) correlation test.
Figure \ref{fig:ppdb-exact-tuned} shows the results for the controlled paraphrase classification setting, where both paraphrase and non-paraphrase pairs have exactly 50\% word overlap. 

The first comparison to note is between original and controlled settings, which allows us to establish the contributions of overlap information as opposed to composition. Comparing between Figure~\ref{fig:bird-tuned} and Figure~\ref{fig:bird-abba-tuned}, it is clear that fine-tuned models still show substantial reduction in correlation when overlap cues are removed. The same goes for Figure \ref{fig:ppdb-exact-tuned} (by comparison to Figure~\ref{fig:ppdb-tuned} of the Appendix)---we see that on the controlled dataset, accuracies hover just above chance-level performance both before and after fine-tuning, compared to over 90\% accuracy on the uncontrolled dataset. This gap in performance between the original and controlled datasets mirrors the findings of \citet{yu2020assessing}, and suggests that even after fine-tuning, the majority of correspondence between model phrase representations and human meaning similarity judgments can be attributed to capturing of word content information rather than phrasal composition.  

The second key comparison is between pre-trained and fine-tuned models within the overlap-controlled settings. While the prior comparison tells us that similarity correspondence is still dominated by word content effects, this second comparison can tell us whether fine-tuning shows at least some boost in meaning composition relative to pre-training. Comparing performance of pre-trained and fine-tuned models in Figure \ref{fig:bird-abba-tuned}, we see that fine-tuning on PAWS-QQP actually slightly degrades correlations at many layers for a majority of models and representation types---with improvements largely restricted to XLM-RoBERTa and XLNet (perhaps notably, mostly in cases where pre-trained correlations are negative). This is despite the fact that models achieve strong validation performance on PAWS-QQP (as shown in Table \ref{tab:paws_acc}), suggesting that learning this task does little to improve composition. We will explore the reasons for this below.

In Figure~\ref{fig:ppdb-exact-tuned}, we see that fine-tuning also does little to improve paraphrase classification accuracies in the controlled setting---though each model shows slight improvement in peak accuracy across layers and representation types (e.g., RoBERTa shows $\sim$3\% increase in peak accuracy with SST tuning, and 2\% with PAWS tuning). 
Even so, the best accuracies across models continue to be only marginally above chance. This, too, fails to provide evidence of any substantial composition improvement resulting from the fine-tuning process.

The story changes slightly when we turn to impacts of SST fine-tuning on correlations in Figure \ref{fig:bird-abba-tuned}. While all correlations remain low after SST fine-tuning, we do see that correlations for BERT, XLM-RoBERTa and XLNet show some non-trivial benefits even in the controlled setting. In particular, SST tuning consistently improves correlation among all representation types in BERT (except for minor degradation in later layers for Head-token), boosting the highest correlation from $\sim$0.2 to $\sim$0.39. Between representation types, the greatest change is in the CLS token, with the most dramatic point of improvement being an abrupt correlation peak for CLS at BERT's fourth layer. We will discuss more below about this localized benefit. 


A final important observation is that fine-tuning on either dataset produces clear degradation in correlations for all representation types in RoBERTa under the controlled setting, by contrast to the general improvements seen for that and other models in the uncontrolled setting. This suggests that at least for that model, fine-tuning encourages retention or enhancement of lexical information, but may degrade compositional phrase information.\footnote{Following \citet{yu2020assessing}, in addition to phrase-only inputs we also try embedding target phrases in sentence contexts. Consistent with the findings of \citet{yu2020assessing}, we see that  presence of context words does boost overall correlation and accuracy, but does not alter the general trends. Moreover, models still show relatively weak performance on controlled tasks even with context available (see Figure~\ref{fig:bird-in-sent-tuned} and Figure~\ref{fig:bird-abba-in-sent-tuned} in the Appendix for details).}


\begin{table}[t!]
\centering
\begin{tabular}{c|c}
\thick
\textbf{Model}        & \textbf{Accuracy(\%)} \\ \thick
BERT    & 80.13        \\ \hline
RoBERTa & 90.81        \\ \hline
DistilBERT & 81.98        \\ \hline
XLM-RoBERTa                           & 91.18                             \\ \hline
XLNet                         & 88.24                             \\ \hline
Linear CLF          & 71.34                             \\ \hline
\end{tabular}
\caption{Accuracy of fine-tuned models on PAWS-QQP dev/test set. Linear CLF is a baseline classifier with relative swapping distance as the only input feature.}
\label{tab:paws_acc}
\end{table}

\begin{figure*}[ht]
    \centering
    \includegraphics[width=1\textwidth]{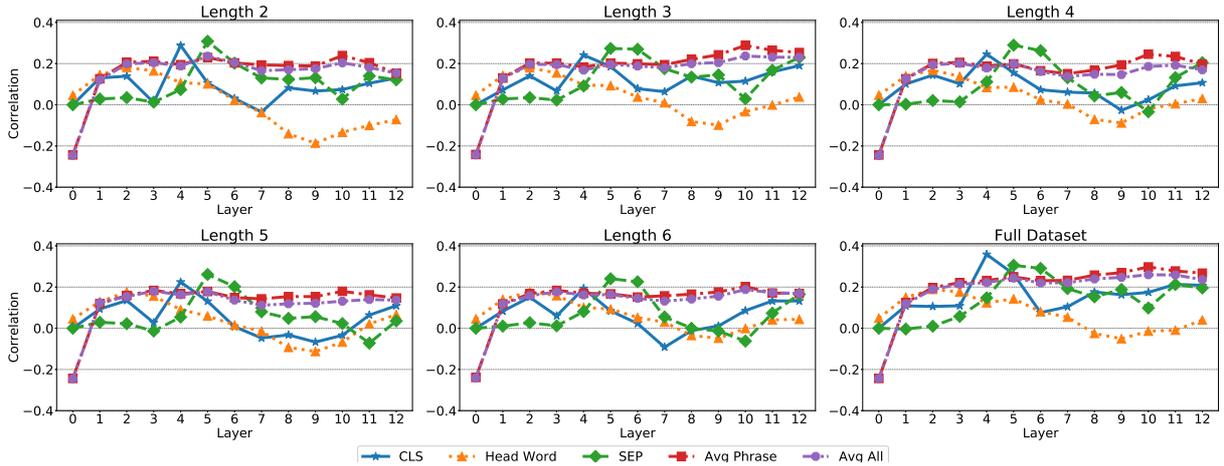}
    \caption{Layer-wise correlation of BERT fine-tuned on phrases of different lengths in SST.}
    \label{fig:abba_various_len}
\end{figure*}

\begin{figure}[t]
    \centering
    \includegraphics[width=0.47\textwidth]{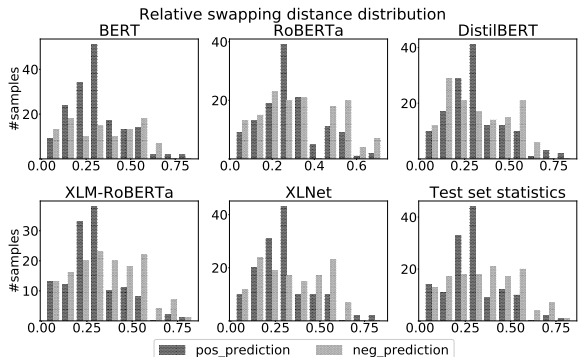}
    \caption{Distribution of positive and negative predictions made by tuned models. Last plot shows the statistics in the PAWS-QQP dev/test set. X-axis corresponds to relative swapping distance; Y-axis shows number of samples in the specific relative swapping distance bin.}
    \label{fig:paws_histo}
\end{figure}

\section{Analyzing impact of fine-tuning}

The presented results suggest that despite compelling reasons to think that fine-tuning on the selected tasks may improve composition of phrase meaning, these models mostly do not exhibit noteworthy benefits from fine-tuning. In particular, fine-tuning on the PAWS-QQP dataset often degrades performance on the controlled datasets taken to be most indicative of compositionality. As for SST, the benefits are minimal, but in localized cases like BERT's CLS token, we do see signs of improved compositionality. In this section, we conduct further analysis on the impacts of fine-tuning, and discuss why tuned models behave as they do.

\subsection{Failure of PAWS-QQP} Table \ref{tab:paws_acc} shows accuracy of fine-tuned models on the dev/test set of PAWS-QQP.\footnote{The performance of BERT in the table is different from previous work mainly due to the fact that models in \citet{zhang2019paws} are tuned on concatenation of QQP and PAWS-QQP datasets rather than PAWS-QQP only.} It is clear that the models are learning to perform well on this dataset, but our results above indicate that this does not translate to improved composition sensitivity. 

\begin{figure*}[ht]
    \centering
    \includegraphics[width=1\textwidth]{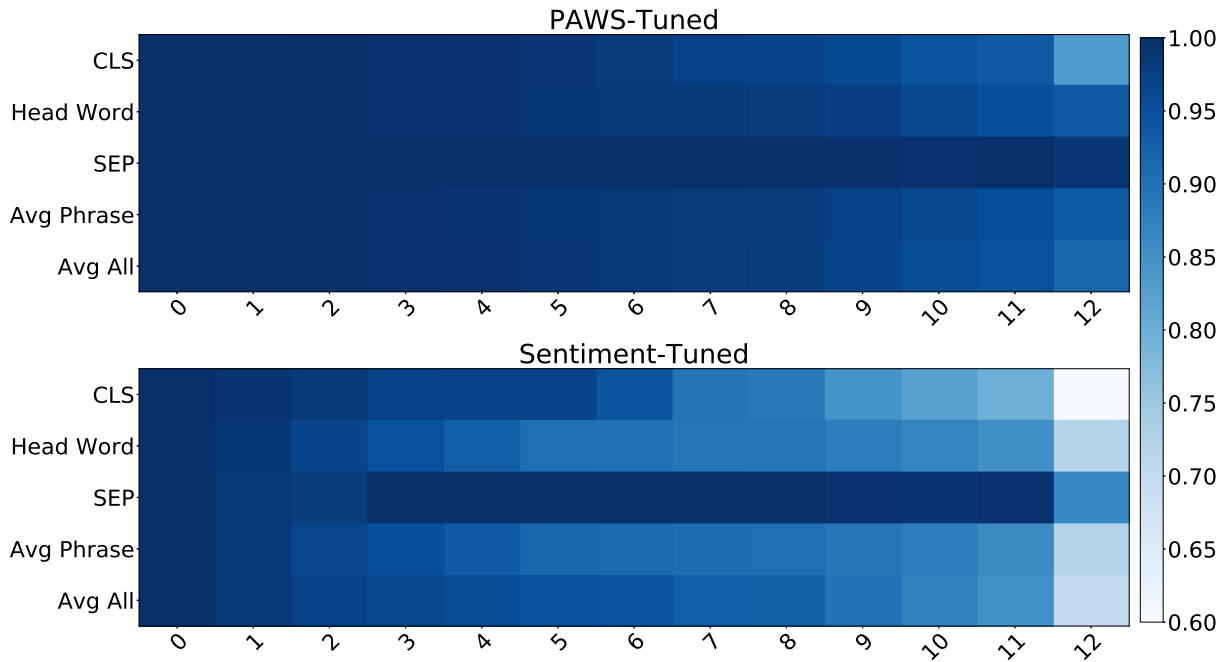}
    \caption{Average layer-wise embedding similarity between fine-tuned and pre-trained BERT. The upper half shows the comparison between PAWS-QQP tuned and pre-trained BERT. And the lower half presents Sentiment Treebank-tuned v.s. pre-trained. Embeddings are evaluated using full BiRD dataset for input.}
    \label{fig:rep_changes}
\end{figure*}

We explore the possibility that this discrepancy may be caused by trivial cues arising during the construction of the dataset, enabling models to infer paraphrase labels without needing to improve their understanding of the meaning of the sentence pair \cite[c.f.,][]{poliak2018hypothesis, gururangan2018annotation}. Sentence pairs in PAWS are generated via word swapping and back translation to ensure high bag-of-words overlap~\cite{zhang2019paws}. We hypothesize that models may be able to achieve high performance in this task based on distance of the word swap alone, without requiring any sophisticated representation of sentence meaning. 

To test this, given a sentence pair $(s_1, s_2)$ with word counts $l_1, l_2$, respectively, we define ``relative swapping distance'' as
\begin{equation*}
    dist_{relative} = \frac{dist_{swap}}{max(l_1, l_2)}
\end{equation*}
where $dist_{swap}$ is defined as the index difference of the first swapping word in $s_1$ and $s_2$. For the example shown in the first row of Table \ref{tab:paws_example}, the first swapping word is ``specific'', with $dist_{swap} = 4$. Note that with this measure we focus on information from one word swap only, while some pairs in PAWS-QQP have multiple swapped words---so in reality, swapping distance information may be even stronger than our results below indicate.

In the last plot of Figure \ref{fig:paws_histo}, we show an association between relative swapping distance and paraphrase labels in the PAWS dev/test set: sentence pairs with small swapping distance tend to be positive samples, while large swapping distance associates with negative labels. The other plots in Figure \ref{fig:paws_histo} show distribution of positive and negative predictions generated by each fine-tuned model with respect to relative swapping distance. We see a similar pattern, with models tending to generate negative labels when swapping distance is larger. 

To verify the viability of this cue, we train a simple linear classifier on PAWS-QQP, with relative swapping distance as the only input feature. The results are reported as ``Linear CLF'' in Table \ref{tab:paws_acc}. Even without access to the content of the sentences, we see that this simple model is able to achieve non-trivial and comparably good classification accuracy on the dev/test set. The strong performance of the linear classifier and the distribution of predictions are consistent with the hypothesis that when we tune on PAWS-QQP, rather than forcing models to learn nuanced meaning in the absence of word overlap cues, we may instead encourage models to focus on lower-level information having little to do with the sentence meaning, further degrading their performance on the composition tasks.

\subsection{Localized impacts of SST}

Fine-tuning on sentiment shows a bit of a different pattern---while it mostly shows only minor changes from pre-training, and the correlations and classification accuracies remain at decidedly low levels on the controlled settings, we do see in certain models some distinctive changes in levels of similarity correlation as a result of tuning on SST. Notably, since these improvement patterns are seen in the similarity correlations but not in the classification accuracies, this suggests that these two tasks are picking up on slightly different aspects of phrasal compositionality. To investigate these effects further, we focus our attention on BERT, which shows the most distinctive improvement in correlations. 

The obvious candidate for the source of the localized SST benefit is the dataset's inclusion of labeled syntactic phrases of various sizes. The benefits seen from SST tuning suggest that this may indeed encourage models to gain some finer-grained sensitivity to compositional impacts of phrase structure (at least those relevant for sentiment). To examine this further, we filter the SST dataset to subsets with phrases of the same length, from 2 to 6 words, and tune pre-trained BERT on each subset. 

Figure~\ref{fig:abba_various_len} shows the correlations for BERT, fine-tuned on each phrase length, on the overlap-controlled BiRD dataset. We see that tuning on the full dataset (mixed phrase lengths) gives the strongest fourth-layer boost in CLS correlation performance---but among the size subsets, a semblance of the fourth-layer CLS peak is seen across phrase lengths, with length-2 training yielding the strongest peak, and length-6 training the smallest. This suggests an amount of size-based specialization---sentiment training on phrases of (or closer to) two words has more positive impact on similarity correlations for our two-word phrases.\footnote{Although subset size can potentially contribute to correlation performance, we find that subset size does not correlate with the performance patterns we observe here. Phrase count of each subset: length 2 - 11,499; length 3 - 11,779; length 4 - 15,050; length 5 - 11,816; length 6 - 9,935.} However, we also see that phrases of other sizes contribute non-trivially to the ultimate correlation improvement observed from training on the full dataset. This is consistent with the notion that training on diverse phrase sizes encourages fine-grained attention to compositionality, while training on phrases of similar size may have slightly more direct benefit. 

\paragraph{Representation changes}  For further comparison of fine-tuning effects between tasks, we analyze changes in BERT representations at each layer before and after the fine-tuning process. Figure~\ref{fig:rep_changes} shows the average layer-wise representation similarity between fine-tuned and pre-trained BERT given identical input. We see substantial differences between tasks in terms of representation changes: while SST fine-tuning produces significant changes across representations and layers, PAWS fine-tuning leaves representations largely unchanged (further supporting the notion that this task can be solved fairly trivially). We also see that after SST tuning, BERT's CLS token shows robust similarity to pre-trained representations until the fifth layer, followed by a rapid drop in similarity. This suggests that the fourth-layer correlation peak may be enabled in part by retention of key information from pre-training, combined with heightened phrase sensitivity from fine-tuning. We leave in-depth exploration of this dynamic for future work.

\section{Discussion} The results of our experiments indicate that despite the promise of PAWS-QQP and SST tasks for improving models' phrasal composition, fine-tuning on these tasks falls far short of resolving the composition weaknesses observed by~\citet{yu2020assessing}. The majority of correspondence with human judgments can still be attributed to word overlap effects---disappearing once overlap is controlled---and improvements on the controlled settings are absent, very small, or highly localized to particular models, layers and representations. This outcome aligns with the increasing body of evidence that NLP datasets often do not require of models the level of linguistic sophistication that we might hope for---and in particular, our identification of a strong spurious cue in the PAWS-QQP dataset adds to the growing number of findings emphasizing that NLP datasets often have artifacts that can inflate performance~\cite{poliak2018hypothesis,gururangan2018annotation,kaushik2018much}.

We do see a ray of promise in the small, localized benefits for certain models from tuning on SST. These improvements do not extend to all models, and are fairly small in the models that do see benefits---but as we discuss above, it appears that training on fine-grained syntactic phrase distinctions may indeed confer some enhancement of compositional meaning in phrase representations---at least when model conditions are amenable. Since sentiment information constitutes only a very limited aspect of phrase meaning, we anticipate that training on fine-grained phrase labels that reflect richer and more diverse meaning information could be a promising direction for promoting composition more robustly in these models. 

\section{Conclusions and future directions}

We have tested effects of fine-tuning on phrase meaning composition in transformer representations. Although we select tasks with promise to address composition weaknesses and reliance on word overlap, we find that representations in the fine-tuned models show little improvement on controlled composition tests, or show only very localized improvements. Follow-up analyses suggest that the PAWS-QQP dataset contains spurious cues that undermine learning of sophisticated meaning properties when training on that task. However, results from SST tuning suggest that training on labeled phrases of various sizes could prove effective for learning composition.
Future work should investigate how model properties interact with fine-tuning to produce improvements in particular models and layers---and should move toward phrase-level training with meaning-rich annotations, which we predict will be a promising direction for improving models' phrase meaning composition.

\section*{Acknowledgments}
We would like to thank three anonymous reviewers for valuable feedback for improving this paper. We also thank members of the University of Chicago CompLing Lab for helpful comments and suggestions on this work. This material is based upon work supported by the National Science Foundation under Award No.~1941160.

\bibliographystyle{acl_natbib}
\bibliography{acl2021}

\newpage
\appendix
\section{Appendix}

\begin{figure*}[ht]
    \centering
    \includegraphics[width=1\textwidth]{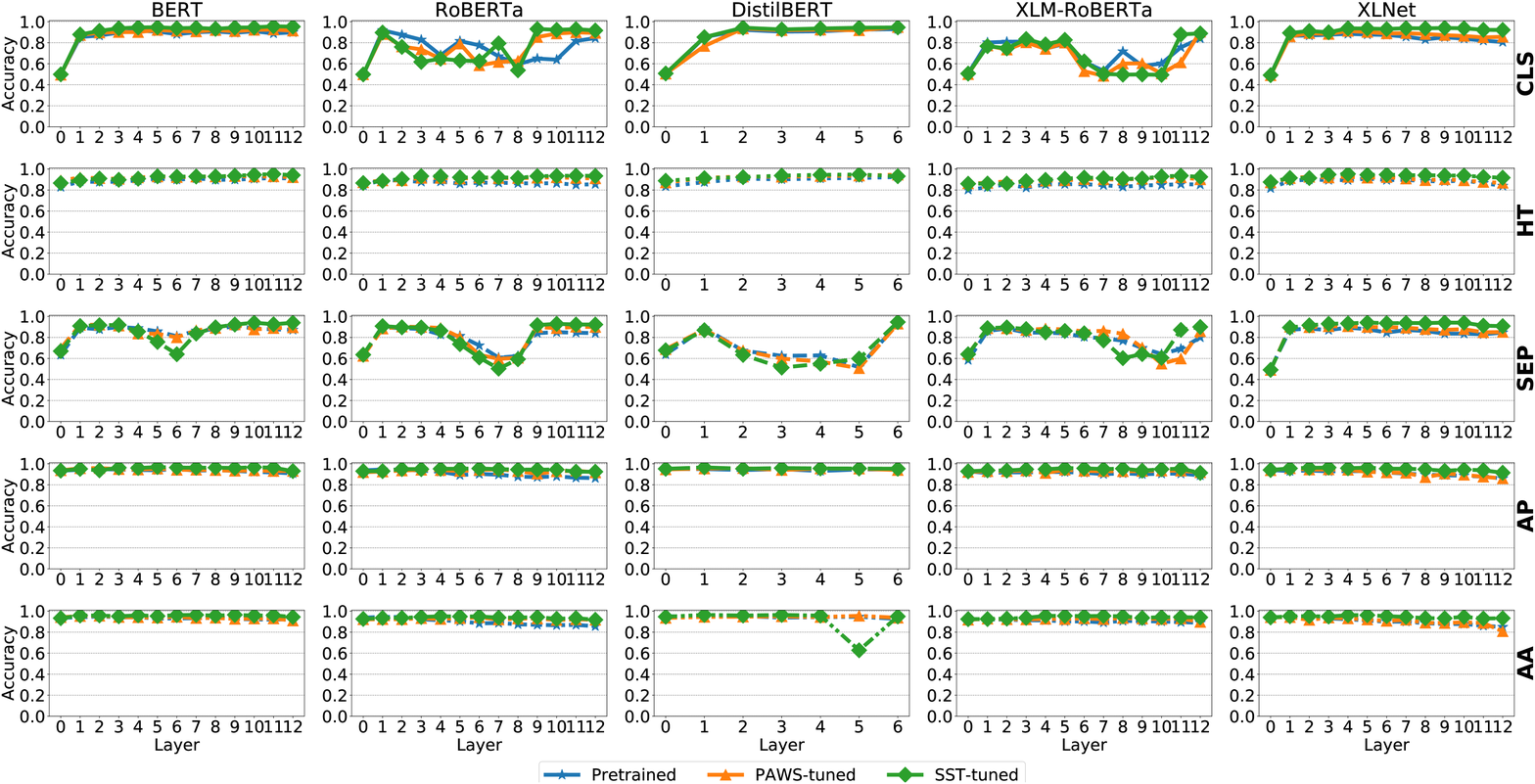}
    \caption{Paraphrase classification accuracy on uncontrolled PPDB dataset, with phrase-only input. Columns correspond to models, and rows correspond to representation types (``HT'' = Head-Token, ``AP'' = Avg-Phrase and ``AA'' = Avg-All). For each model and representation type, the corresponding subplot shows accuracies for pre-trained, PAWS-tuned and SST-tuned settings, respectively. For each subplot, X-axis corresponds to layer index, and Y-axis corresponds to accuracy value. Layer 0 corresponds to input embeddings passed to the model.}
    \label{fig:ppdb-tuned}
\end{figure*}

\begin{figure*}[t]
    \centering
    \includegraphics[width=1\textwidth]{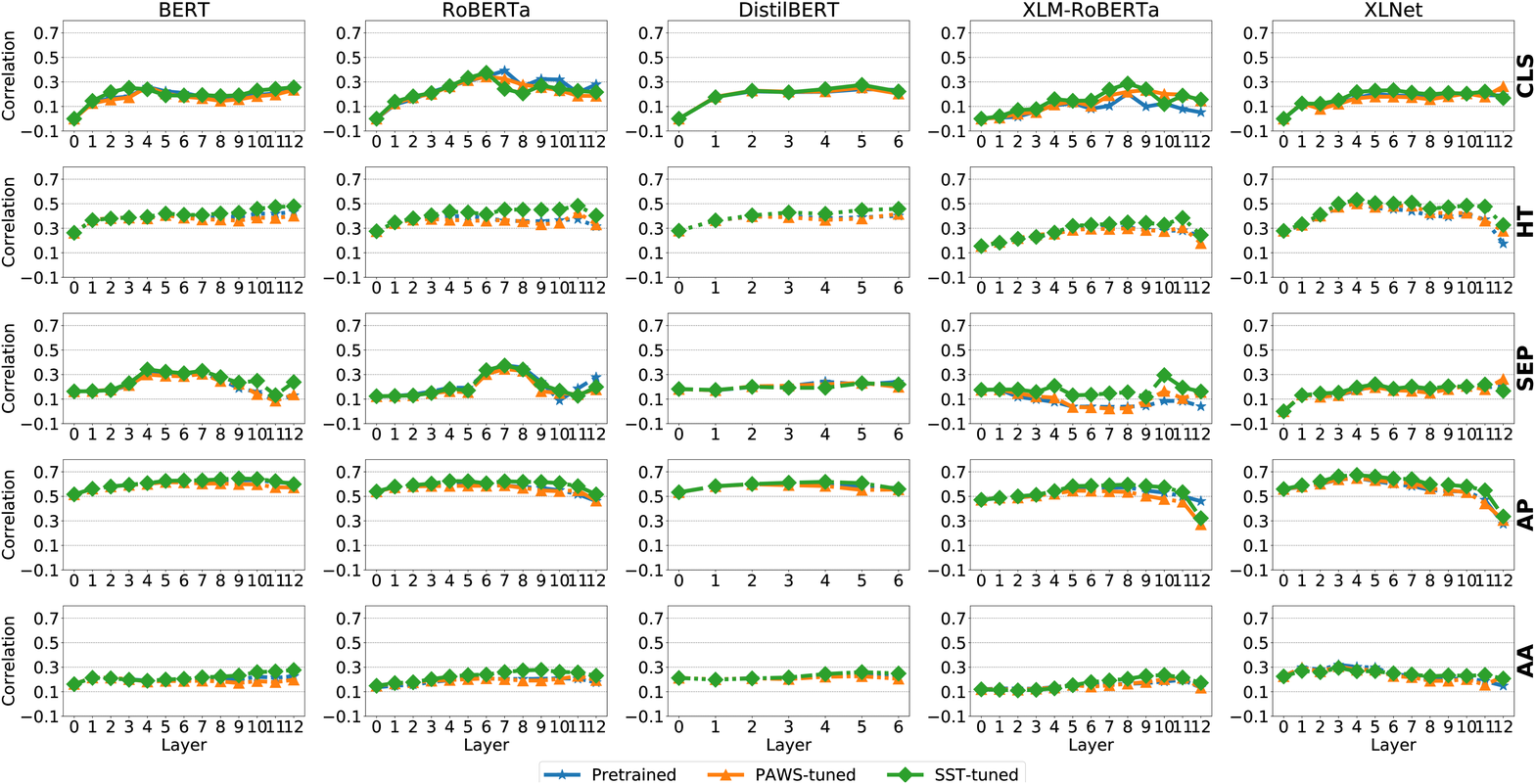}
    \caption{Similarity correlation on full BiRD dataset with phrases embedded in context sentence (context-available input).}
    \label{fig:bird-in-sent-tuned}
\end{figure*}

\begin{figure*}[h]
    \centering
    \includegraphics[width=1\textwidth]{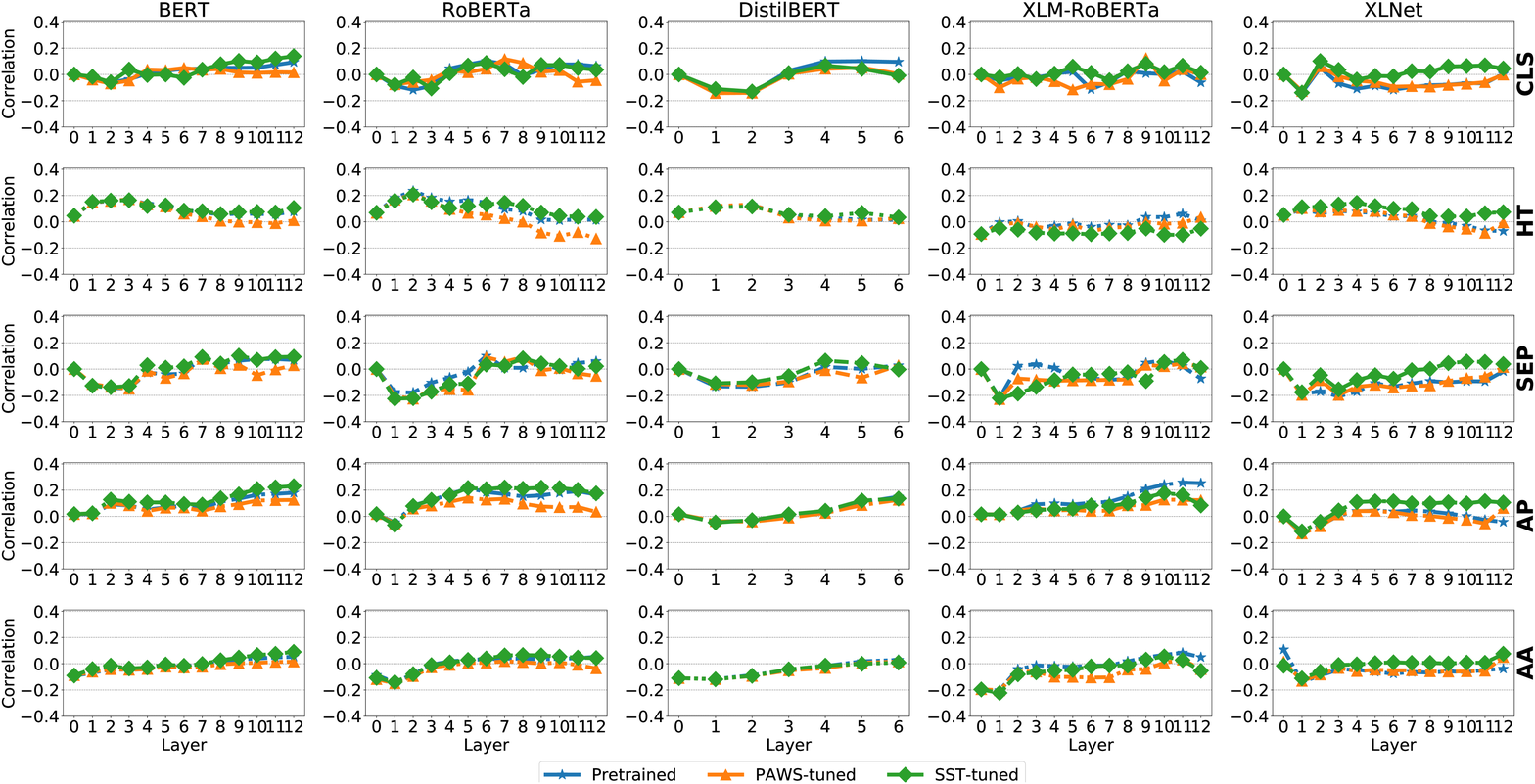}
    \caption{Similarity correlation on controlled BiRD dataset (AB-BA setting) with phrases embedded in context sentence (context-available input).}
    \label{fig:bird-abba-in-sent-tuned}
\end{figure*}
\end{document}